\documentclass[sigconf]{acmart}

\AtBeginDocument{%
  \providecommand\BibTeX{{%
    \normalfont B\kern-0.5em{\scshape i\kern-0.25em b}\kern-0.8em\TeX}}}

\copyrightyear{2023}
\acmYear{2023}
\setcopyright{acmlicensed}

\acmConference[SIGIR '23] {Proceedings of the 46th International ACM SIGIR Conference on Research and Development in Information Retrieval}{July 23--27, 2023}{Taipei, Taiwan.}
\acmBooktitle{Proceedings of the 46th International ACM SIGIR Conference on Research and Development in Information Retrieval (SIGIR '23), July 23--27, 2023, Taipei, Taiwan}
\acmPrice{15.00}
\acmISBN{978-1-4503-9408-6/23/07}
\acmDOI{10.1145/3539618.3591675}

\usepackage{amsmath}
\usepackage{amsfonts}
\usepackage{url}
\newcommand{\donotdisplay}[1]{}

\graphicspath{{Figures/}}
\usepackage{siunitx}
\usepackage{multirow}
\usepackage{colortbl,etoolbox,siunitx,xcolor}
\robustify\bfseries
\newcommand{\maxf}[1]{\bfseries #1}
\newcolumntype{P}[1]{>{\centering\arraybackslash}p{#1}}
\usepackage{mathtools}
\def\multiset#1#2{\ensuremath{\left(\kern-.3em\left(\genfrac{}{}{0pt}{}{#1}{#2}\right)\kern-.3em\right)}}

\begin{document}

\title[Dynamic Mixed Membership Stochastic Block Model for Weighted Labeled Networks]{Dynamic Mixed Membership Stochastic Block Model for Weighted Labeled Networks}

\author{Ga\"el Poux-M\'edard}
\affiliation{%
  \institution{Université de Lyon, Lyon 2, UR 3083}
  \streetaddress{5 avenue Pierre Mendès France}
  \city{Bron}
  \country{France}}
\email{gael.poux-medard@univ-lyon2.fr}

\author{Julien Velcin}
\affiliation{%
  \institution{Université de Lyon, Lyon 2, UR 3083}
  \streetaddress{5 avenue Pierre Mendès France}
  \city{Bron}
  \country{France}}
\email{julien.velcin@univ-lyon2.fr}

\author{Sabine Loudcher}
\affiliation{%
  \institution{Université de Lyon, Lyon 2, UR 3083}
  \streetaddress{5 avenue Pierre Mendès France}
  \city{Bron}
  \country{France}}
\email{sabine.loudcher@univ-lyon2.fr}

\renewcommand{\shortauthors}{Poux-M\'edard et al.}

\begin{abstract}
Most real-world networks evolve over time. Existing literature proposes models for dynamic networks that are either unlabeled or assumed to have a single membership structure.
On the other hand, a new family of Mixed Membership Stochastic Block Models (MMSBM) allows to model static labeled networks under the assumption of mixed-membership clustering. In this work, we propose to extend this later class of models to infer dynamic labeled networks under a mixed membership assumption. Our approach takes the form of a temporal prior on the model's parameters. It relies on the single assumption that dynamics are not abrupt. We show that our method significantly differs from existing approaches, and allows to model more complex systems --dynamic labeled networks. We demonstrate the robustness of our method with several experiments on both synthetic and real-world datasets.
A key interest of our approach is that it needs very few training data to yield good results. The performance gain under challenging conditions broadens the variety of possible applications of automated learning tools --as in social sciences, which comprise many fields where small datasets are a major obstacle to the introduction of machine learning methods.
\end{abstract}




\received{20 February 2007}
\received[revised]{12 March 2009}
\received[accepted]{5 June 2009}

\maketitle

\section{Introduction}
\begin{figure}
    \centering
    \includegraphics[width=\columnwidth]{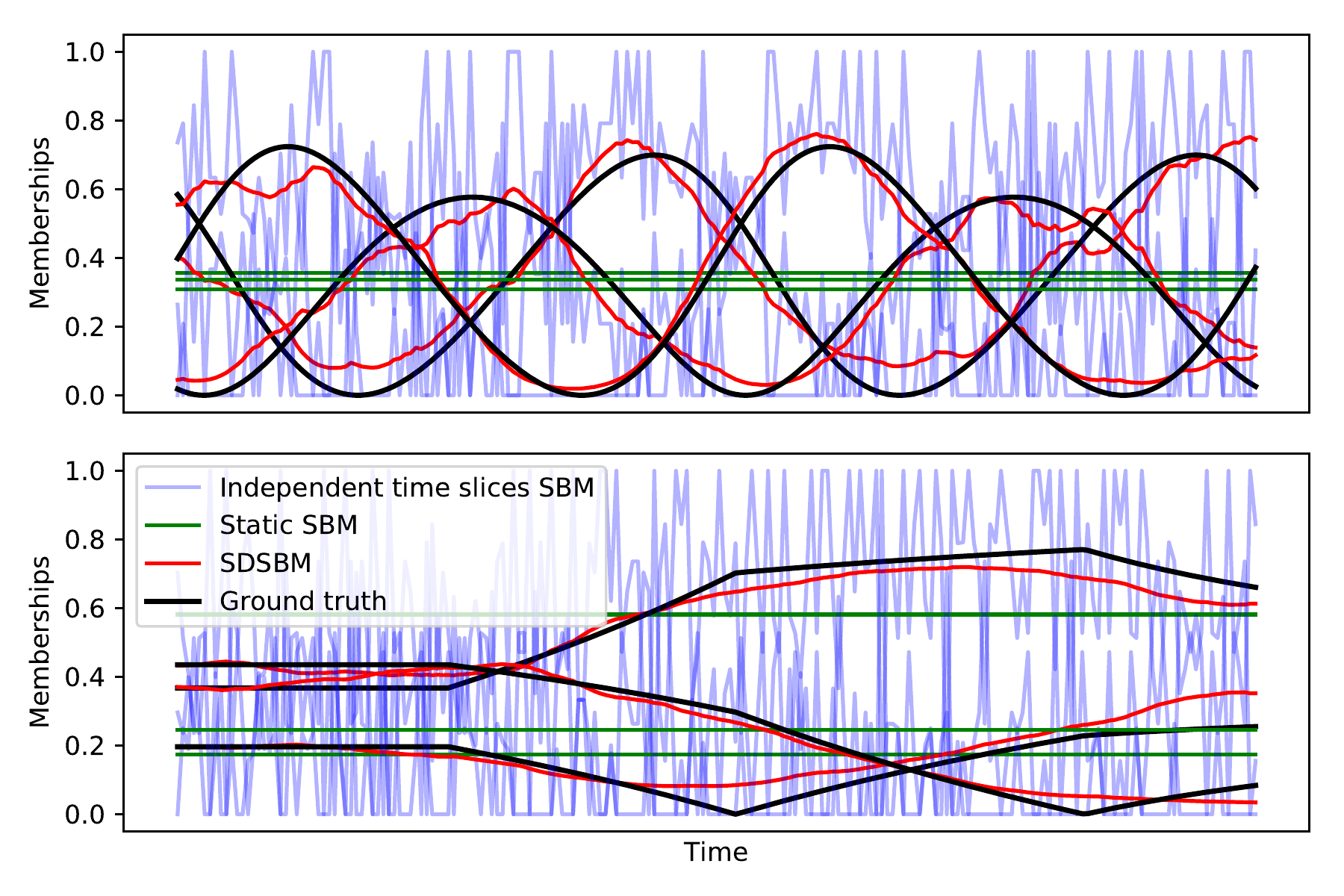}
    \caption{\textbf{Users' attachment to groups can vary over time} --- A music listener could cyclically prefer Rock, Jazz or Pop music (top), or listen to either of these without any specific pattern (bottom). For 200 epochs containing only 5 observations each, our approach (in red) infers any smooth dynamic membership pattern, and does it more accurately than static models (in green \cite{Antonia2016AccurateAndScalableRS}) and models that consider each time slice independently (in blue \cite{Tarres2019TMBM}).
    }
    \label{fig-illustration}
\end{figure}

Dynamic networks are powerful tools to visualize and model interactions between different entities that can evolve over time. The network's nodes represent the interacting entities, and ties between these nodes represent an interaction. In many real-world situations, ties strength can vary over time --on music streaming websites for instance, users' affinity with various musical genres vary over time \cite{Kumar2019jodie,RothVillermet2021MusicConsoSpotify} --see Fig.~\ref{fig-illustration} as an illustration. The network is said dynamic.

Now, every interaction does not have the same significance. A music listener might like both Rock and Jazz, but favor one over the other. This user's ties to musical genres do not have the same intensity; each tie is associated to a number, representing the strength of the interaction. The network is said to be dynamic and weighted. 

Finally, simple dynamic weighted networks might not be fit to grasp the complexity of a given situation. A music listener may have different opinions on musical genres; they can like it, dislike it, be bored of it, prefer to listen to some only in the morning, at night, etc. Each of these relations can be represented by their own tie in the network, each associated to their own strength. The network is said to be dynamic, weighted and labeled.

Such networks are high-dimensional objects, whose direct inference is a difficult problem. Several ways to achieve this task in a static setting have been proposed; the Stochastic Block Models (SBM) family is one of the most popular approaches \cite{Holland1983SBM,Guimera2013DrugdrugSBM,CoboLopez2018SocialDilemma}. The underlying assumption is that some sets of nodes behave in a similar way regarding other sets of nodes. They can be grouped in a cluster; instead of modeling every edge for every node in a network, only the edges between clusters are modeled, making the inference task much more tractable. Each cluster is associated to labeled edges, and each node is associated to a cluster. A variant of SBM that allows more expressive power is called the Mixed-Membership SBM (MMSBM), where each node can belong to several clusters with different proportions \cite{Yuchung1987,Airoldi2008MMSBM,Antonia2016AccurateAndScalableRS,Tarres2019TMBM}. The inferred network is then weighted and labeled, but not dynamic. A major advantage of the SBM family is that they yield readily interpretable results (interpretable clusters), unlike most existing neural-network approaches for labeled networks \cite{Fan2021InterpretabilityNeuralNet}.

The goal of this paper is to infer networks that are dynamic, weighted and labeled by using a mixed-membership SBM. Our contribution consists in the extension of a broad class of MMSBM by using a temporal Dirichlet prior.
We first conduct a careful review of dynamic network inference literature, and we show that no prior work addresses the task at hand. Although some previous works tackle similar problems, they are not fit for inferring dynamic, weighted \textit{and} labeled networks. Besides, our approach is conceptually much simpler than those proposed in the literature. Last but not least, our dynamic extension is readily plugable into most existing MMSBM for labeled and weighted networks.

We develop an EM optimization algorithm that scales linearly with the size of the dataset, demonstrate the effectiveness of our approach on both synthetic and real-world datasets, and further detail a possible application scenario.

\section{Background}
\subsection{Notations}
We consider a network of $I$ nodes and $O$ labels. All the clustering models discussed in this section can be represented by a restricted set of parameters: a matrix $\theta(t) \in \mathbb{R}^{I \times K}$ accounts for the membership of each of $I$ nodes to each of $K$ possible clusters, and a block-interaction matrix $p(t) \in \mathbb{R}^{K \times K \times O}$ represents the probability for each of $K$ clusters to yield each of $O$ labels. Both $\theta (t)$ and $p (t)$ can vary over time; in the following, this temporal dependence is implicit ($x(t) := x$) unless specified otherwise. The network is said to be unlabeled when $O=1$, and binary (as opposed to weighted) if an edge can only exist or not exist.

\subsection{Dynamic unlabeled networks - Single-membership}
Single-membership SBMs consider a membership matrix such as $\theta \in \{0;1\}^{I \times K}$: each membership vector equals $1$ for one cluster, and $0$ everywhere else --``hard'' clustering in the literature.
In \cite{Xu2014DynamicSBM,Xu2014DynamicSBM2}, the authors proposed to model a binary unlabeled dynamic network using a label-switching inference method, optimized by a Sequential Monte Carlo algorithm \cite{Jin2021ReviewSBMs}. Both the membership and the interaction matrices can vary over time, thus supposing two independent underlying Markov processes \cite{Jin2021ReviewSBMs}.

In \cite{Yang2010DetectingCA,Tang2014}, the authors propose to model a binary dynamic and unlabeled network. The cluster interaction matrix $p$ can vary over time while keeping the memberships $\theta$ static. The entries of $p$ are drawn from a Dirichlet distribution and expressed as a Chinese Restaurant Process. This process converges to a Dirichlet distribution, and allows to infer a potentially infinite number of clusters. This model is therefore non-parametric and inferred using an MCMC algorithm. 

A conceptually novel way of modeling dynamic unlabeled networks under the single membership assumption has been proposed in \cite{Matias2017DynSBM,Matias2018DynSBM}. The authors propose to model the cluster interaction and membership matrices as Poisson processes, that explicitly represent the continuous temporal dependency without slicing the dataset into episodes. The method allows to infer varying membership \textit{and} interaction matrices for dynamic binary or weighted networks, but their results have shown that allowing both to vary simultaneously leads to identifiability and label switching issues \cite{Funka2019ReviewSBM}. This conclusion seems reasonable, given none of these SBM algorithms can reach a global optimum of the likelihood function. A model where both $\theta$ and $p$ can vary over time is more likely to get stuck into a local optimum.

Finally, we mention the existence of SBM variants that account for dynamic degree-correction \cite{Wilson2019ModelingDynDegCorr} or that enforce a scale-free characteristic \cite{Wu2019ScaleFreedSBM}. 

All these methods consider unlabeled networks, and consider a hard clustering which does not allow for as much expressive power as the Mixed-Membership approaches.

\subsection{Dynamic unlabeled networks - Mixed-membership}
Mixed-membership SBMs consider a membership matrix such as $\theta \in \mathbb{R}^{I \times K}$, where each membership vector $\theta_i$ has a $L_1$ norm of $1$. Literature also refer to it as ``soft'' clustering.

Similar to \cite{Yang2010DetectingCA,Tang2014}, a method for inferring dynamical binary unlabeled networks has been proposed in \cite{Fan2015DynInfMMSBM}. The membership vector of each piece of information is drawn from a Chinese Restaurant Process (CRP) according to the number of times a node has already been associated to each cluster before. The process yields a distribution over an infinity of available clusters. The formulation as a CRP arises naturally, because the prior on membership vectors is typically a Dirichlet distribution --a CRP naturally converges to a draw from a Dirichlet distribution \cite{Arratia1992}. The block-interaction matrix $p$ does not vary over time. The article shows a complexity analysis that suggest the methods runs with a complexity of $\mathcal{O}(N^2)$ which makes it unfit for large-scale real-world applications.

The work the most closely related to ours is \cite{Xing2010dMMSBM}. This seminal work proposed the dMMSB as a way to model dynamic binary unlabeled networks using a variational algorithm \cite{Lee2019ReviewSBMs}. To do so, the authors modify the original MMSBM \cite{Airoldi2008MMSBM} to consider a logistic normal distribution as a prior on the membership vectors $\theta_i$. This choice allows to model correlations between membership vectors' evolution \cite{Ahmed2007LogisticNormal}. The membership vectors are then embedded in a state space model, that is a space where one can define a linear transition between two time points for a given variable. The authors define such trajectory for the membership vectors as a linear function of the previous time point. The trajectory is estimated and smoothed using a Kalman Filter. This approach is the most closely related to ours, as it models the temporal dependency using a prior distribution over memberships, noted $P(\theta)$. \cite{Xing2010dMMSBM} has been extended to consider $P(\theta)$ as a logistic normal mixture prior \cite{Ho2011dM3SB}, which improves its expressive power. 

However, this model and related extensions are not fit for the task at hand. First, it considers unlabeled and binary networks \cite{Lee2019ReviewSBMs}, and extension to labeled and weighted networks, if possible, is not trivial. The proposed optimization algorithm requires a loop until convergence at each EM iteration, making it unable to handle large datasets. Besides, the clusters interaction matrix $p$ must remain static for the approach to work, which we alleviate here. And most importantly, is assumes a linear transition between time steps in the state space, while we do not assume any kernel function in our proposed approach. 

\subsection{Static labeled networks - Mixed-membership}
Recent years saw a rise of Bayesian methods for inferring static weighted labeled networks using MMSBM variants \cite{Antonia2016AccurateAndScalableRS,Tarres2019TMBM,Poux2021MMSBMMrBanks,Poux2021IMMSBM,Poux2022SIMSBM}. Here, nodes are associated a type $a$; each type of node has its own layer in a multipartite network, and its associated set of available clusters through a membership matrix $\theta^a$. Nodes from one layer belong to every cluster corresponding to this layer with mixed proportions. The interaction between clusters of all layers then yields a distribution over $O$ possible output labels. This results in a static MMSBM for labeled networks. Our contribution extends all of \cite{Antonia2016AccurateAndScalableRS,Tarres2019TMBM,Poux2021MMSBMMrBanks,Poux2021IMMSBM,Poux2022SIMSBM} to the dynamic case. These works will be further detailed in Section~\ref{section-model}.

Note that in \cite{Tarres2019TMBM}, the authors consider a temporal slicing of the data and consider each slice as independent from the others; a time slice is considered as a node in a tripartite network. We will compare our approach to this modeling later. 

We use these works as a base model, that we couple to a dynamic prior distribution on parameters. Our work focuses on making the prior probability of both $\theta$ and $p$ time-dependent, in order to model these parameters dynamics. We provide a ready-to-use temporal plug-in for each of the works presented in this section. It applies to dynamical, weighted and labeled networks in a mixed-membership context, and inference is conducted with a scalable variational EM algorithm.

\section{Dynamic labeled MMSBM}
\label{section-model}
\subsection{Base model}
For clarity, we choose to consider the \textbf{simplest form} of a labeled MMSBM: each of $I$ nodes is associated to each of $K$ clusters in mixed proportions, and each of $K$ clusters is in turn associated to $O$ labels (SIMSBM(1) in \cite{Poux2022SIMSBM}). As shown in \cite{Poux2022SIMSBM}, our demonstration on SIMSBM(1) trivially extends to \cite{Antonia2016AccurateAndScalableRS} (SIMSBM(1,1)) \cite{Tarres2019TMBM} (SIMSBM(2,1)) and \cite{Poux2021IMMSBM} (SIMSBM(2)) --and in general any model formulated as SIMSBM(x,y,...). 

We consider a set of $I$ nodes that can be associated to $O$ possible labels on a discrete time interval, or epoch, written $t$. We assume that, at each time step, each of $I$ nodes belongs to a mixture of $K$ available clusters, each of which are in turn yield a probability distribution over $O$ labels. The membership of each of $I$ nodes to each of the $K$ possible clusters at time $t$ is encoded in the membership matrix $\theta^{(t)} \in \mathbb{R}^{I \times K}$. One vector $\theta_i^{(t)}$ represents the probability that $i$ belongs to any of the $K$ clusters at time $t$, and is normalized as:
\begin{equation}
    \label{eq-normtheta}
    \sum_{k \in K} \theta_{i,k}^{(t)} = 1 \ \forall i, t
\end{equation}
The probability for each of $K$ clusters to be associated to each of $O$ labels at time $t$ is encoded in the matrix $p^{(t)} \in \mathbb{R}^{K \times O}$. An entry $p_k^{(t)}(o)$ represents the probability that cluster $k$ is associated to label $o$ at time $t$, and thus is normalized as:
\begin{equation}
    \label{eq-normp}
    \sum_{o \in O} p_k^{(t)}(o) = 1 \ \forall k, t
\end{equation}
Finally, the probability that a node $i$ is associated to label $o$ at time $t$ (i.e. the probability of an edge between $i$ and $o$ at time $t$) is written:
\begin{equation}
    P(i \rightarrow o \vert t) = \sum_{k \in K} \theta_{i,k}^{(t)} p_k^{(t)}(o)
\end{equation}
Given a set $R^{\circ}$ of observed triplets $(i,o,t)$, the model's posterior distribution can be written \cite{Antonia2016AccurateAndScalableRS,Poux2022SIMSBM}:
\begin{align}
    \label{eq-L}
    P(\theta, p \vert R^{\circ}) &\propto P(R^{\circ} \vert \theta, p) \prod_t P(\theta^{(t)})P(p^{(t)}) \\ 
    = \prod_{(i,o,t) \in R^{\circ}} &\sum_{k \in K} \theta_{i,k}^{(t)} p_k^{(t)}(o) \prod_t \left( \prod_i P(\theta_i^{(t)}) \prod_k P(p_k^{(t)}) \right) \notag 
\end{align}
Now, before we describe the optimization procedure, we must choose the priors $P(\theta^{(t)})$ and $P(p^{(t)})$.

\subsection{Simple Dynamic SBM (SDSBM) prior}
We formulate the prior distribution over $\theta^{(t)}$ and $p^{(t)}$ under a single assumption: the parameters at a given time do not vary abruptly at small time scales. It means an entry $\theta^{(t_1)}_{i,k}$ should not differ significantly from $\theta^{(t_2)}_{i,k}$ for every $t_2$ close enough to $t_1$. The entries close to a reference time are called \textbf{temporal neighbours}.

\begin{figure}
    \centering
    \includegraphics[width=\columnwidth]{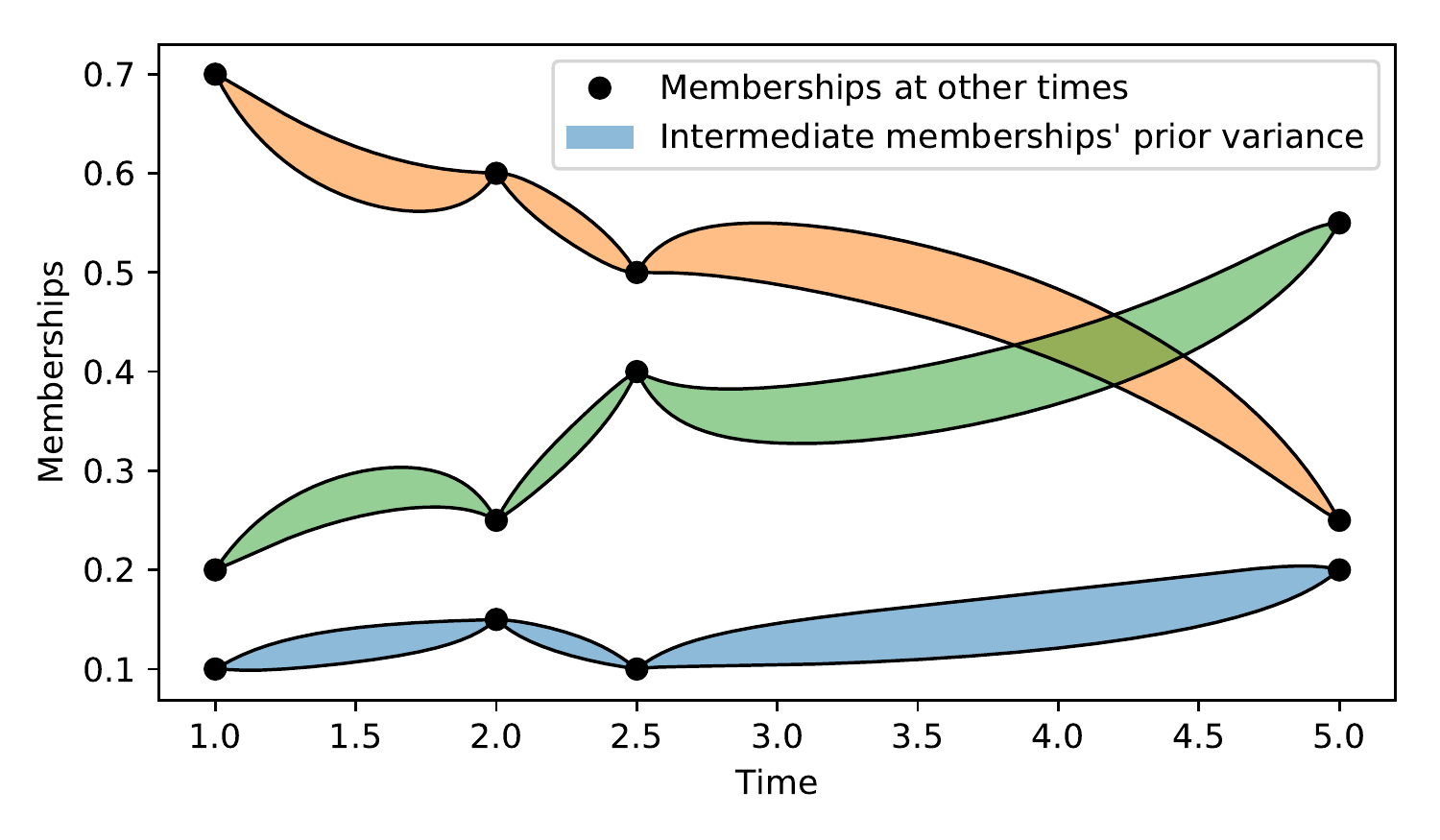}
    \caption{\textbf{Prior probability's variance on memberships at all times according to the temporal neighbourhood} --- Variance of the prior over a membership entry (filled curves, we represented 3 such entries in different colors as illustration) as a function of time, given some temporal neighbours (black dots). This illustration considers an averaging kernel as $\kappa(t,t') = \frac{1}{\vert t-t' \vert}$. When inferring a parameter $x^{(t)}$ at a time $t$, the variance of its prior probability $P(x^{(t)})$ depends on $t$ relative to the temporal neighbours. Here for instance, the variance is null at $t=2$ because $\kappa(t,t')$ diverges, and so does $\alpha^{(t)}$, hence the variance collapsing to 0.}
    \label{fig-illustration-kernel}
\end{figure}

Our \textit{a priori} knowledge on each entry $\theta_i^{(t)}$ and $p_k^{(t)}$ is that they should not differ significantly from their temporal neighbours. This is a fundamental difference with \citep{Xing2010dMMSBM}, where the next parameters values are estimated using from a Kalman Filter that only considers the previous time step. Moreover, the authors assume a linear transition function, while we do not make such hypothesis. An illustration of the proposed approach is given Fig.~\ref{fig-illustration-kernel}, where the prior probability of a membership over time depends on its temporal neighbours.

\subsubsection{Dirichlet distribution}
Since each entry $\theta_i^{(t)}$ and $p_k^{(t)}$ is normalized to 1, we consider a Dirichlet distribution as a prior, which naturally yields normalized vectors such that $\sum_n x_n = 1$. It reads:
\begin{equation}
    \label{eq-dir}
    Dir(x \vert \alpha) = \frac{1}{B(\alpha)}\prod_n x_n^{\alpha_n-1}
\end{equation}
where $B(\cdot)$ is the multivariate beta function. In Eq.~\ref{eq-dir}, the vector $\alpha$ is called the concentration parameter and must be provided to the model. This parameter defines the mode and the variance of the Dirichlet distribution.

\begin{figure}
    \centering
    \includegraphics[width=0.9\columnwidth]{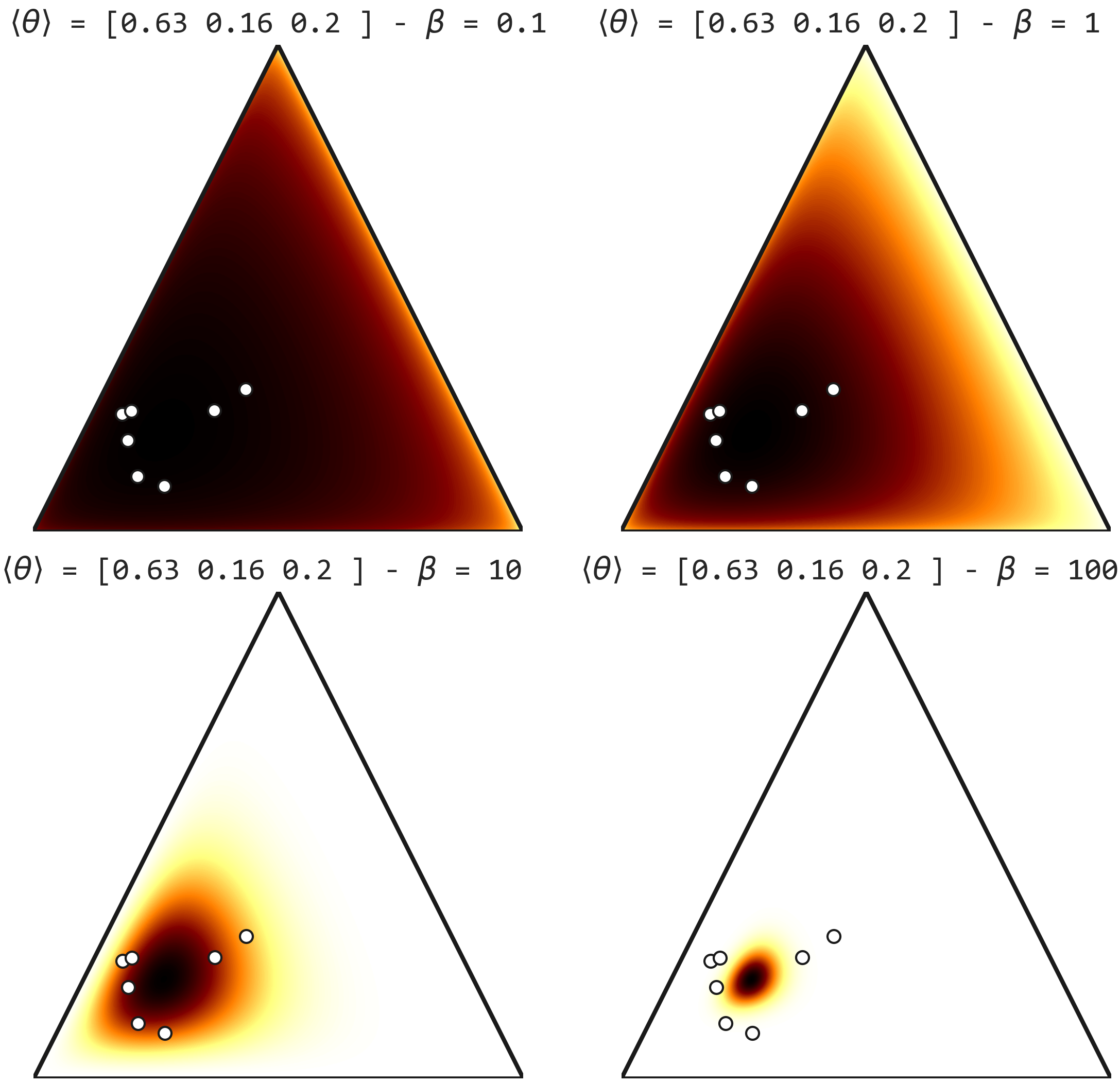}
    \caption{\textbf{Prior probability on a membership vector for various values of $\beta$ according to temporal neighbourhood} --- Darker means higher probability. Projected on a simplex tri-space (each of 3 axes ranges from 0 to 1). The white dots represent the temporal neighbours of the considered 3D vector. Their average is given as $\langle \theta \rangle$ using a uniform weight function $\kappa(t,t')$ for illustration purpose. $\beta$ controls the variable's prior variance around its neighbours.}
    \label{fig-dirichlet}
\end{figure}

We consider a concentration parameter as $\alpha = 1 + \beta \alpha_0$, so that when $\beta=0$ we recover a uniform prior over the simplex, and $\beta \geq 0$ so that the prior has a unique mode.
The most frequent value drawn from Eq.\ref{eq-dir} (or mode) is $\mathbb{M}(x_n) = \frac{\alpha_n-1}{\sum_n' (\alpha_n' - 1)} = \frac{\alpha_{0,n}}{\sum_n' \alpha_{0,n}}$. We recover a uniform prior for $\beta=0$; the variance vanishes with $\beta \gg 1$ as $\frac{1}{\beta}$. The effect of various values of $\beta$ on the prior distribution is illustrated in Fig.~\ref{fig-dirichlet}.

\subsubsection{Simple Dynamic prior's mode}
Our single constraint is that $\theta_i^{(t)}$ and $p_k^{(t)}$ do not vary abruptly over small time scales. To do so, we define their prior probability's mode with respect to close temporal neighbours. The hyper-parameter $\beta$ controls the variance of the prior --that is, how much our constraint should impact the inference procedure. We express the Simple Dynamic prior parameters for $\theta_i^{(t)}$ as:
\begin{equation}
    \label{eq-prior}
    \alpha_{i,k}^{(t,\theta)} = 1+ \beta \underbrace{\left(\frac{\sum_{t' \neq t} \kappa(t,t') \theta_{i,k}^{(t')}}{\sum_{t' \neq t} \kappa(t,t')}\right)}_{\langle\theta_{i,k}^{(t)}\rangle}
\end{equation}
where $\kappa(t,t')$ is a weight function, and $\alpha^{(t,\theta)}$ corresponds to the concentration parameter for $\theta$ at time $t$. In following experiments, we define the weight function as $\kappa(t,t') = \frac{N_{t'}}{\vert t-t' \vert}$, where $N_{t'}$ is the number of observations made at time $t'$. This way, temporal neighbours' influence decrease as the inverse of temporal distance. 
We illustrate the influence of this particular kernel function on the prior probability on membership at all times in Fig.~\ref{fig-illustration-kernel}. 
In particular we see that with this expression, the prior probability variance goes to 0 when the considered time it very close to a temporal neighbour.

The mode of the prior is then the average value of its the temporal neighbours weighted by $\kappa(t,t')$, noted $\langle\theta_{i,k}^{(t)}\rangle$. Note that this holds because $\sum_k \langle\theta_{i,k}^{(t)}\rangle = 1 \ \forall i,t$. Besides, the prior variance is a decreasing function of $\beta$; when $\beta=0$ the prior is uniform over the simplex, and when $\beta \rightarrow \infty$ the variance goes to $0$, as illustrated Fig.~\ref{fig-dirichlet}. The same reasoning holds for $p_k^{(t)}$, with prior parameters $\alpha_{k,o}^{(t,p)} = 1 + \beta \langle p_{k}^{(t)}(o)\rangle$, as well as for any other parameter in \cite{Antonia2016AccurateAndScalableRS,Tarres2019TMBM,Poux2021IMMSBM,Poux2022SIMSBM}.

\subsubsection{Priors expression}
Finally, we give the final log-priors on $\theta_i^{(t)}$ and $p_k^{(t)}$:
\begin{align}
    \label{eq-priors}
    P(\theta_i^{(t)} \vert \{\theta_{i,k}^{(t')}\}_{t' \neq t}) &\propto \prod_k {\theta_{i,k}^{(t)}}^{\beta \langle\theta_{i,k}^{(t)}\rangle}\\
    P(p_k^{(t)}(o) \vert \{p_{k}^{(t')}(o)\}_{t' \neq t}) &\propto \prod_o {p_{k}^{(t)}(o)}^{\beta \langle p_{k}^{(t)}(o)\rangle} \notag 
\end{align}
We omitted the normalisation factor for clarity --it does not influence the inference procedure.

\subsection{Inference}
\subsubsection{E step}
We develop an EM inference procedure for maximizing the log-posterior distribution defined Eq.\ref{eq-L}. The expectation step computes the expected probability of a latent variable (here a cluster $k$) being chosen given each entry of $R^{\circ}$. Since such latent variables do not appear in the priors expressions, the expectation step remains unchanged by the introduction of the Simple Dynamic Priors; in general, prior distributions do not intervene in the computation of the expectation step \cite{Bishop2006}. The E step for such labeled networks has already been discussed on several occasions \cite{Antonia2016AccurateAndScalableRS,Tarres2019TMBM,Poux2021IMMSBM}. It can be derived as:
\begin{align}
    \label{eq-L}
    \log P(\theta, p \vert R^{\circ}) &\propto \log P(R^{\circ} \vert \theta, p) \prod_t\prod_i P(\theta_i^{(t)})\prod_k P(p_k^{(t)}) \notag \\ 
    = &\sum_{(i,o,t) \in R^{\circ}} \log \sum_{k \in K} \theta_{i,k}^{(t)} p_k^{(t)}(o) \\
    &+ \sum_t \sum_i \log P(\theta_i^{(t)}) \sum_k \log P(p_k^{(t)}) \notag \\
    \geq &\sum_{(i,o,t) \in R^{\circ}}\sum_{k \in K} \omega_{i,o}^{(t)}(k) \log \frac{\theta_{i,k}^{(t)} p_k^{(t)}(o)}{\omega_{i,o}^{(t)}(k)} \notag \\
    &+ \sum_t \sum_i \log P(\theta_i^{(t)}) \sum_k \log P(p_k^{(t)}) \notag
\end{align}
In Eq.\ref{eq-L}, we introduced a proposal distribution $\omega_{i,o}^{(t)}(k)$, that represents the probability of one cluster allocation $k$ given the observation $(i,o,t)$. The last line followed from Jensen's inequality assuming that $\sum_k \omega_{i,o}^{(t)}(k) = 1$. We notice that Jensen's inequality holds as an equality when:
\begin{equation}
    \label{eq-omega}
    \omega_{i,o}^{(t)}(k) = \frac{\theta_{i,k}^{(t)} p_k^{(t)}(o)}{\sum_{k'} \theta_{i,k'}^{(t)} p_{k'}^{(t)}(o)}
\end{equation}
which provides us with the expectation of the latent variable $k$ given an observation $(i,o,t) \in R^{\circ}$, written $\omega_{i,o}^{(t)}(k)$.

Using this expression, we can rewrite the log-likelihood $\log P(R^{\circ} \vert \theta, p)$ as \cite{Bishop2006,Antonia2016AccurateAndScalableRS,Poux2022SIMSBM}:
\begin{equation}
    \label{eq-newlogl}
    \log P(R^{\circ} \vert \theta, p) = \sum_{(i,o,t) \in R^{\circ}} \sum_{k \in K}  \omega_{i,o}^{(t)}(k) \log \frac{\theta_{i,k}^{(t)} p_k^{(t)}(o)}{\omega_{i,o}^{(t)}(k)}
\end{equation}

\subsubsection{M step}
Taking back the first line of Eq.\ref{eq-L} and substituting with Eq.\ref{eq-priors} and Eq.\ref{eq-newlogl}, we get an unconstrained expression of the posterior distribution. We introduce Lagrange multipliers to account for the constraints of Eq.\ref{eq-normtheta} ($\phi_i^{(t)}$) and Eq.\ref{eq-normp} ($\psi_i^{(t)}$), and finally compute the maximization equations with respect to the model's parameters. Starting with the membership matrix entries $\theta_{i,k}^{(t)}$:
\begin{align}
    &\frac{\partial \left(\log P(\theta, p \vert R^{\circ}) - \sum_{i',t'} \phi_{i'}^{(t')}(\sum_{k'}\theta_{i',k'}^{(t')}-1)\right)}{\partial \theta_{i,k}^{(t)}} = 0 \notag \\
    &\leftrightarrow \sum_{o \in \partial(i,t)} \frac{\omega_{i,o}^{(t)}(k)}{\theta_{i,k}^{(t)}} + \frac{\beta\langle\theta_{i,k}^{(t)}\rangle}{\theta_{i,k}^{(t)}} - \phi_{i}^{(t)} = 0 \notag \\
    &\leftrightarrow \sum_{o \in \partial(i,t)} \omega_{i,o}^{(t)}(k) + \beta\langle\theta_{i,k}^{(t)}\rangle = \phi_{i}^{(t)}\theta_{i,k}^{(t)} \notag \\
    &\leftrightarrow \sum_{o \in \partial(i,t)} \underbrace{\sum_k \omega_{i,o}^{(t)}(k)}_{=1 \text{ (Eq.\ref{eq-omega})}} + \beta\underbrace{\sum_k \langle\theta_{i,k}^{(t)}\rangle}_{=1 \text{ (Eq.\ref{eq-normtheta})}} = \phi_{i}^{(t)} \notag \\
    &\leftrightarrow \frac{\sum_{o \in \partial(i,t)} \omega_{i,o}^{(t)}(k) + \beta\langle\theta_{i,k}^{(t)}\rangle}{N_{i,t}+\beta} = \theta_{i,k}^{(t)}
\end{align}
where $\partial(i,t)=\{ o \vert (i, \cdot, t) \in R^{\circ} \}$ is the subset of labels associated to both $i$ and $t$, and $N_{i,t} = \vert \partial(i,t) \vert$ is the size of this set. Note that for $\beta = 0$ we recover the M-step of standard static MMSBM models \cite{Antonia2016AccurateAndScalableRS,Tarres2019TMBM,Poux2021IMMSBM,Poux2022SIMSBM}.

The derivation of the M-step for the entries $p_{k}^{(t)}(o)$ is identical and yields
:
\begin{align}
    p_{k}^{(t)}(o) = \frac{\sum_{(i,t) \in \partial o} \omega_{i,o}^{(t)}(k) + \beta\langle p_{k}^{(t)}(o)\rangle}{\sum_{(i,o,t) \in R^{\circ}} \omega_{i,o}^{(t)}(k)+\beta}
\end{align}

\subsection{Discussion}
\subsubsection{Plug-in for existing models}
Using SDSBM (for Simple Dynamic SBM) as a temporal extension in existing MMSBM models requires very few changes. In \cite{Antonia2016AccurateAndScalableRS,Tarres2019TMBM,Poux2021IMMSBM,Poux2022SIMSBM}, its introduction boils down to adding a term $\beta\langle x \rangle$ to the numerator of maximization equations, and the corresponding normalizing term $\beta$ to the denominator. This way, our approach is ready-to-use to make these models for for modelling dynamic networks.

\subsubsection{Flexible dynamic modeling}
The prior allows to consider that some parameters are dynamic and that others are not. For instance, when several membership matrices are involved, as in \cite{Antonia2016AccurateAndScalableRS,Tarres2019TMBM,Poux2022SIMSBM}), setting $\beta=0$ for some makes them time-invariant (or universal). The Simple Dynamic prior also allows to choose whether the block-interaction tensor $p$ is dynamic. 
Moreover, $\beta$ does not have to be the same for every membership matrix, or even every entry $i$ of each of them. Finally, $\beta$ itself can vary over time. 
To summarize, $\beta$ allows to control the temporal scale over which parameters may vary. This allows to jointly model universal parameters ($\beta=0$) and dynamical ones ($\beta \neq 0$).

\subsubsection{Tunable temporal dependence}
Finally, the choice of the averaging kernel function $\kappa(t,t')$ is important. It allows to choose the range over which the inference of a variable should rely on its temporal neighbours. A formulation as the inverse of time difference seems relevant: the weight of a neighbour appearing at a time $\delta t$ later should diverge as $\delta t \rightarrow 0$, so that continuity is ensured. Besides, one could control the smoothness of the curve with respect to time by tuning the weight function as $\kappa(t,t')=\frac{N_{t'}}{\vert t-t' \vert^{a}}$ where $a=1,2,...$ for instance, where $N_{t'}$ is the number of observations in the time slice $t'$.
Overall, the Simple Dynamic prior works by inferring the variables using both microscopic and mesoscopic temporal scales. If a time slice $t$ has few observations but some of its neighbours have a greater number of them, learning the parameters at $t$ is helped mostly by the population of its closest ($\frac{1}{\vert \Delta t \vert}$) and most populated ($N_{t'}$) neighbours, and less influenced by further and less populated epochs. This is what is illustrated in Fig.~\ref{fig-illustration-kernel}.

\section{Experiments}
\begin{figure}
    \centering
    \includegraphics[width=\columnwidth]{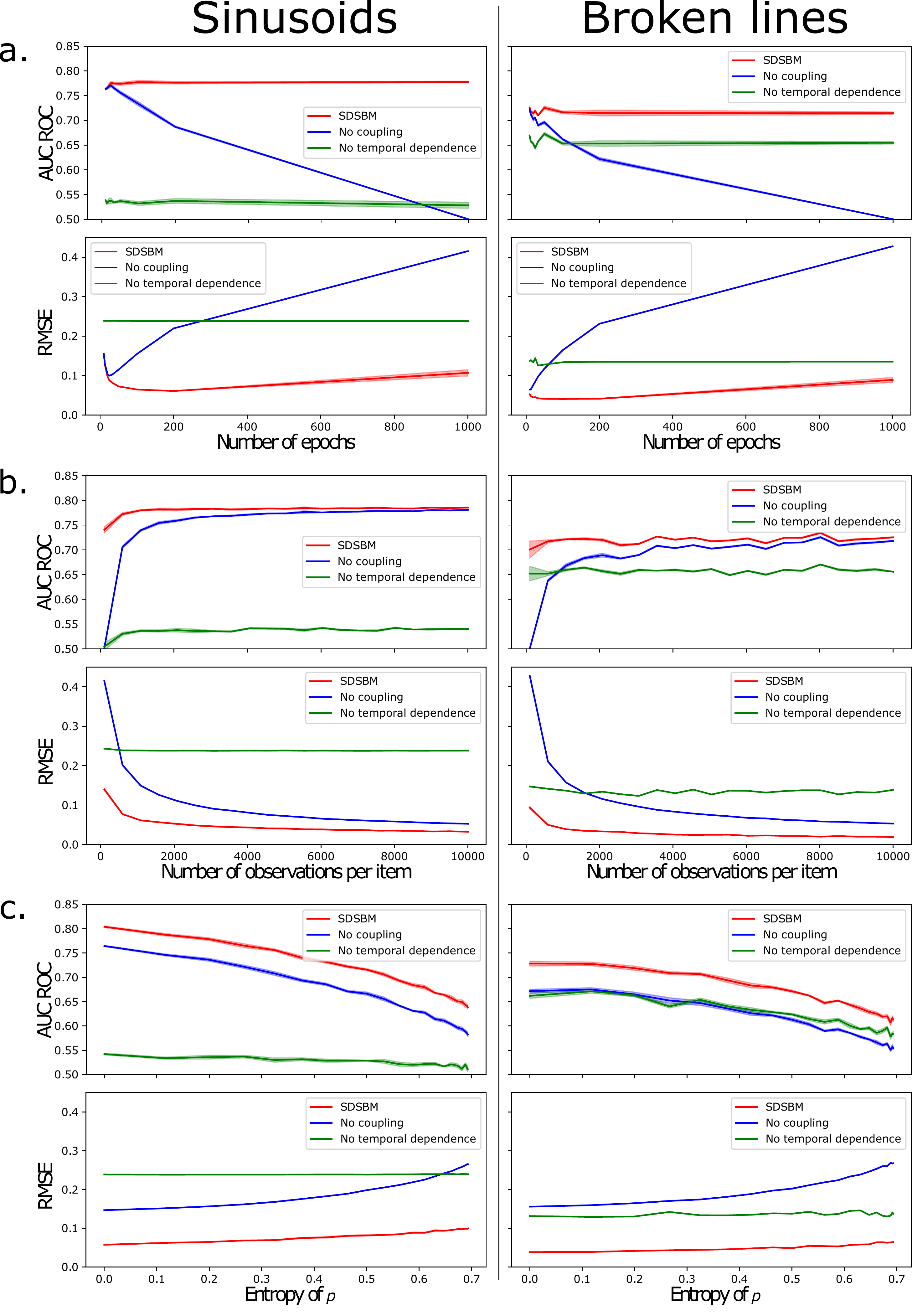}
    \caption{\textbf{Results on synthetic data} --- \textbf{(a.)} SDSBM retrieves the correct dynamic memberships and is little influenced by the data slicing. \textbf{(b.)} SDSBM works well on tiny datasets. \textbf{(c.)} SDSBM retrieves dynamic memberships in challenging situations.}
    \label{fig-resSynth}
\end{figure}
\subsection{Synthetic data}
In this section, we develop several situations in which our method (abbreviated SDSBM for Simple Dynamic MMSBM) could be useful\footnote{Code and data available at \url{https://anonymous.4open.science/r/SDSBM-26DC}}.
Experiments are run for $I=100$, $K=3$ and $O=3$, which are standard testing parameters in the literature of dynamic networks inference \cite{Fan2015DynInfMMSBM,Matias2017DynSBM}. We choose to infer a dynamic membership matrix $\theta^{(t)}$ and to provide a universal block-interaction matrix $p\ \forall t$. Note that the model yields good performances when $p$ also has to be inferred, but due to identificability and label-switching issues raised in \cite{Matias2017DynSBM}, there is not unbiased way to assess the correctness of the inferred memberships values. 
Additional experiments showed that inferring jointly $p$ and $\theta$ yields the same prediction accuracy.
The expression we use for this matrix $p$ is given Eq.\ref{eq-xpp}. We systematically test two variation patterns for $\theta$: a sinusoidal pattern (Fig.~\ref{fig-illustration}-top) and a broken-line pattern (Fig.~\ref{fig-illustration}-bottom). Each pattern is generated with different coefficient for each item; the memberships still sum to 1 at all times. 

\begin{equation}
    \label{eq-xpp}
    p = \begin{bmatrix}
        1-s & s & 0\\
        0 & 1-s & s\\
        s & 0 & 1-s
    \end{bmatrix}
\end{equation}

To the best of our knowledge, the only attempt to model dynamic parameters in labeled weighted and dynamic networks using a MMSBM is \cite{Tarres2019TMBM}. In this work, each epoch is modeled independently from the others. We refer to this baseline as the ``No coupling'' or ``\textbf{NC}'' baseline. For reference, we also compare to a baseline that does not consider the temporal dimension and infers a single universal value for each variable (\textbf{SIMSBM(1)}) \cite{Poux2022SIMSBM}. It has already been assessed that MMSBMs typically outperform state of the art models (NMF, TF, KNN, etc.) in \cite{Antonia2016AccurateAndScalableRS} and \cite{Poux2022SIMSBM}. Such study is not repeated in the present article, for being out of the scope of our demonstration.
We recall that \textbf{we purposely chose the simplest form of MMSBM} for clarity; the reported results may therefore be poor in absolute, but underline an improvement over equally simple models (NC and SIMSBM(1)) brought by our Simple Dynamic prior approach.

We systematically perform a 5-folds cross validation. The model is trained on 80\% of the data, $\beta$ is tuned using 10\% as a validation set, and the model is evaluated on the 10\% left. We choose as metrics the AUC-ROC and the RMSE on the real values of $\theta$ (black line in Fig.~\ref{fig-illustration}). The procedure is repeated 5 times; the error bars reported in the experimental results represent the standard error over these folds.

\subsubsection{SDSBM unveils complex temporal patterns}
In Fig.~\ref{fig-resSynth}a., we consider 1~000 observations for each item $i \in I$ and vary the number of epochs from 10 to 1~000. In the expression of $p$, $s$ is set to 0.05. For both the sinusoidal and line-broken memberships, the model shows better predictive performances (in terms of AUC ROC) than the proposed baselines. Interestingly, the SDSBM performances remain stable as the number of epochs increases unlike the NC baseline, which means it alleviates a bias of the temporal modeling proposed in \cite{Tarres2019TMBM}. 
The RMSE with respect to the true parameters remains low over the whole range of tested number of epochs. The RMSE increases as the number of epochs grows because the number of parameters to estimate increases with it; this makes the inference more subject to local variations, which in turns mechanically increases the RMSE. 
Overall, SDSBM recovers dynamic variations of the membership vectors with a good performance; a sample of the inferred dynamic memberships is shown Fig.~\ref{fig-illustration}.

\subsubsection{SDSBM works with little data}
A major problem that arises when considering temporal data is the scarcity of observations, because slicing implies reducing the number of observations in each slice. This concern largely arises in social sciences, where data retrieval cannot be automated and requires tedious human labor. Here, we demonstrate that our method works in challenging conditions, when data is scarse. In Fig.~\ref{fig-resSynth}b., we vary the number of observations available for each item from 100 to 10.000, distributed over 100 epochs. Thus, in the most challenging situation, there is only one observation per epoch used to determine $I$ dynamic memberships over 3 clusters. In the expression of $p$, $s$ is set to 0.05. We see Fig.~\ref{fig-resSynth}b. that for both patterns, the predictive power of SDSBM remains high in such conditions. 
Moreover, the RMSE on the true dynamic memberships in this case is fairly low, and decreases rapidly as the number of observations increases. This is due to SDSBM linking the time slices together during the training phase: the smoothing constraint makes it so that every time slices indirectly benefits from the training data of its neighbours. 

When the number of observations is high, the ``no coupling'' baseline \cite{Tarres2019TMBM} reaches the performances of SDSBM. This is because as the number of observations in each slice goes to infinity, the models need less to rely on temporal neighbours. However, even for 10.000 observations per item (100 observations per epoch), SDSBM yields better results. As an illustration, the results in Fig.~\ref{fig-illustration} have been obtained using only 5 observations per epoch.

\subsubsection{SDSBM handles highly stochastic interaction patterns}
Finally, we control the deterministic character of the block-interaction matrix $p$ by varying $s$. We express such character as the mean entropy of $p$ $\langle S(p) \rangle$ with respect to its possible outputs: $\langle S(p) \rangle = \frac{1}{K}\sum_{k \in K}\sum_o p_k(o)\log p_k(o)$. The maximum entropy for the proposed expression of $p$ is reached for $s=0.5$. We consider 1~000 observations spread over 100 epochs. We show in Fig.~\ref{fig-resSynth}c. that the predictive performance of all three methods drops as the entropy increases. This is expected, as observations are generated from the true model with a higher variance; each observation becomes less informative about the generative process as $s$ grows. 
However, the RMSE on the real parameters inferred using SDSBM remains low at the maximum entropy, meaning the model recovers the correct membership parameters.

\subsection{Real-world data}
\subsubsection{Experimental setup}
Finally, we demonstrate the validity of our approach on real-world data to argue for its usefulness and scalability. SDSBM builds on previous works on labeled MMSBM and shares the same linear complexity $\mathcal{O}(\vert R^{\circ}\vert)$ with $\vert R^{\circ}\vert$ the size of the dataset \cite{Antonia2016AccurateAndScalableRS}. For our experiments, we consider the recent and documented datasets from \cite{Kumar2019jodie}. The \textbf{Reddit} dataset (10.000 users, 984 subreddits they contribute to, $\sim$670k observations), the \textbf{LastFm} dataset (980 users, 1000 songs users listened to, $\sim$1.3M observations) and the Wikipedia (\textbf{Wiki}) dataset (8227 users, 1000 pages users edited, $\sim$157k observations). The goal is to predict over time what subreddit a user will contribute to, what songs they will listen to, and what Wikipedia pages they will edit. The Reddit and Wikipedia datasets contain 1 month of data; we slice them in 1 day long temporal intervals. The LastFm dataset spans over approximately 5 years; we slice it into periods of 3 days each. In addition, we build an additional dataset (\textbf{Epi}) about historical epigraphy data \footnote{Clauss-Slaby repository, \url{http://www.manfredclauss.de/fr/index.html}}. The dataset is made of 117.000 latin inscriptions comprising one or several of 18 social status (slave, soldier, senator, etc.) and its location as one of 62 possible regions, along with an estimated datation spanning from 100BC to 400AD. The goal is to guess the region where a status has been found, with respect to time. The goal is to recover statuses diffusion in roman territories. We slice this dataset in epochs of one year each. 

\begin{table}
	\caption{\textbf{Numerical results on real-world datasets} --- Metrics abbreviations stand for the area under the ROC curve (ROC), the Average Precision (AP), the Normalized Coverage Error (NCE). Metrics for models stand for Simple Dynamic SDM (SDSBM), No Coupling baseline (NC) and the classical static mixed membership SBM (SIMSBM(1)). Overall, our approach allows for a higher predictive power.}
	\label{table-res}
	\centering
	\begin{tabular}{|l|l|S|S|S|S}
		\cline{1-5}
		& & \text{ROC} & \text{AP} & \text{NCE} \\ 

		\cline{1-5}
		\multirow{3}{*}{\rotatebox[origin=c]{90}{\footnotesize \text{\textbf{Epi}}}}
		& \underline{SDSBM} & \maxf{\num{ 0.9025 +- 0.0011 } }& \maxf{\num{ 0.37 +- 0.0017 } }& \maxf{\num{ 0.1151 +- 0.0011 } }\\
		& NC & \num{ 0.842 +- 0.0022 } & \num{ 0.3435 +- 0.0036 } & \num{ 0.1582 +- 0.0019 } \\
		& SIMSBM(1) & \num{ 0.8597 +- 0.0012 } & \num{ 0.2141 +- 0.0016 } & \num{ 0.1451 +- 0.0013 } \\

		\cline{1-5}
		\multirow{3}{*}{\rotatebox[origin=c]{90}{\footnotesize \text{\textbf{Lastfm}}}}
		& \underline{SDSBM} & \maxf{\num{ 0.8942 +- 0.0008 } }& \maxf{\num{ 0.0168 +- 0.0001 } }& \maxf{\num{ 0.1284 +- 0.0011 } }\\
		& NC & \num{ 0.8393 +- 0.0005 } & \num{ 0.0157 +- 0.0002 } & \num{ 0.1785 +- 0.0007 } \\
		& SIMSBM(1) & \num{ 0.8647 +- 0.0005 } & \num{ 0.0115 +- 0.0002 } & \num{ 0.1493 +- 0.0004 } \\

		\cline{1-5}
		\multirow{3}{*}{\rotatebox[origin=c]{90}{\footnotesize \text{\textbf{Wiki}}}}
		& \underline{SDSBM} & \maxf{\num{ 0.9759 +- 0.0002 } }& \maxf{\num{ 0.0659 +- 0.0009 } }& \maxf{\num{ 0.0459 +- 0.0003 } }\\
		& NC & \num{ 0.9092 +- 0.0007 } & \num{ 0.0608 +- 0.001 } & \num{ 0.1195 +- 0.0008 } \\
		& SIMSBM(1) & \num{ 0.9576 +- 0.0007 } & \num{ 0.0622 +- 0.0004 } & \num{ 0.0565 +- 0.0008 } \\

		\cline{1-5}
		\multirow{3}{*}{\rotatebox[origin=c]{90}{\footnotesize \text{\textbf{Reddit}}}}
		& \underline{SDSBM} & \maxf{\num{ 0.9803 +- 0.0003 } }& \maxf{\num{ 0.4295 +- 0.0054 } }& \maxf{\num{ 0.0312 +- 0.0003 } }\\
		& NC & \num{ 0.8508 +- 0.0005 } & \num{ 0.3598 +- 0.0017 } & \num{ 0.1846 +- 0.0007 } \\
		& SIMSBM(1) & \num{ 0.9798 +- 0.0002 } & \maxf{\num{ 0.4269 +- 0.004 } }& \num{ 0.0322 +- 0.0003 } \\
		\cline{1-5}
	\end{tabular}
\end{table}

Evaluation is again conducted using a 5-folds cross validation with 80\% of training data, 10\% of validation data and 10\% of testing data for each fold. For each pair $(i, o_{true})$ in the test set, we query the probability for every output $o$ given $i$ and build the confusion matrix by comparing them to $o_{true}$.
In Table~\ref{table-res}, we present the results of our method compared to the proposed baselines for various metrics: AUC-ROC (\textbf{ROC}), Average Precision (\textbf{AP}) and Normalized Coverage Error (\textbf{NCE}). The first two metrics evaluate how well models assign probabilities to observations, and the latter evaluates the order in which possible outputs are ranked. 
\begin{figure*}
    \centering
    \includegraphics[width=\textwidth]{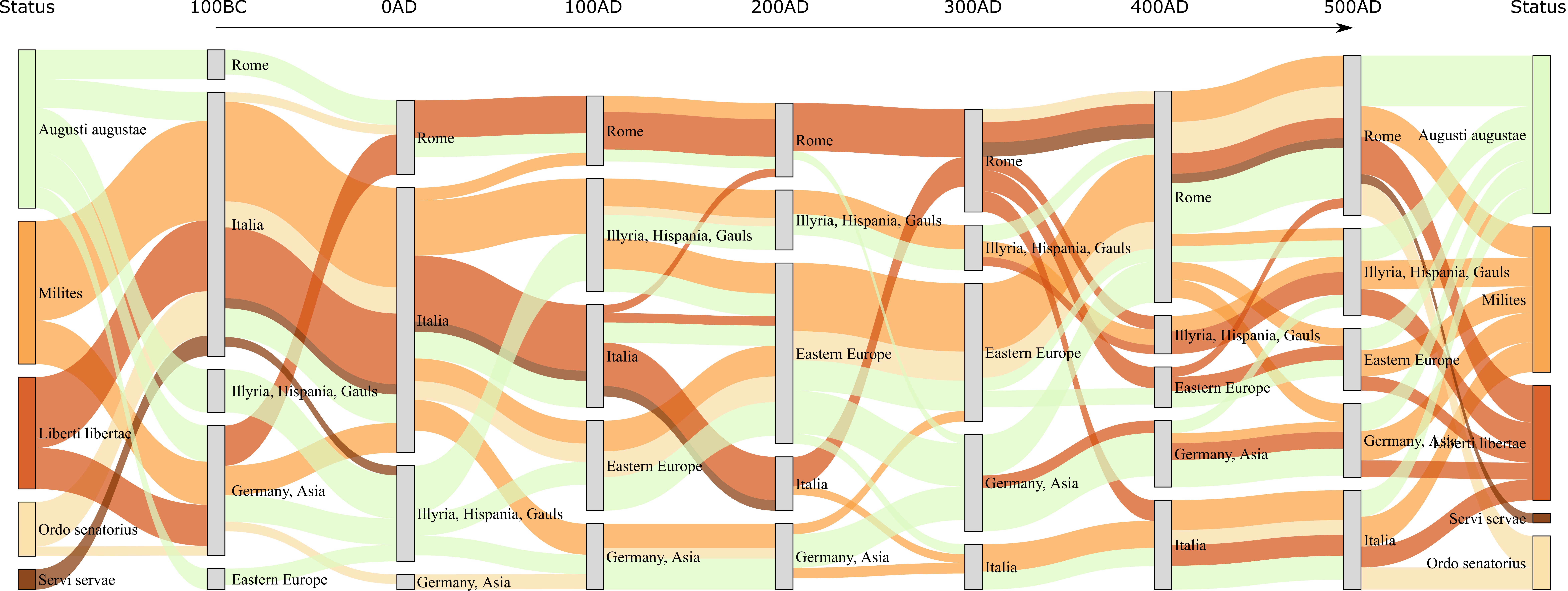}
    \caption{\textbf{Geographic evolution of status distribution from latin graves (100BC - 500AD)} --- We applied the SDSBM to the Epigraphy dataset. We recall that our goal is to predict a roman region (e.g. Illyria, Hispania, etc.) given a status (e.g. Slave, Senator, etc.) and a year. We plot the temporal evolution of statuses membership to the five manually labeled clusters (in gray). For clarity, we removed small membership transfers from the data, which explains why the total cluster's population may vary from one time to another. This plot allows to visualize some global historical trends about the evolution of the Roman Empire (e.g. 3rd century crisis, spread of military presence in Europe, Italy demilitarization, etc.).}
    \label{fig-epigraphy}
\end{figure*}

\subsubsection{Results}
Overall, we see that our method exhibits a greater predictive power, except for the Reddit dataset where the static SBM performs as well as SDSBM. We explain this by the lack of significant temporal variation over the considered interval. This could be expected, since the dataset comprises roughly 80\% of repeated actions \cite{Kumar2019jodie}, meaning that users do not significantly explore new communities over a month. This result shows that SDSBM also works well in the static case. On the other datasets, SDSBM performs better often by a large margin, especially for the ROC-AUC, meaning that SDSBM is efficient at distinguishing classes from each other. We also checked that our initial assumption on the smooth temporal variation of the parameters holds. For every real-world dataset, the absolute average change in membership between two consecutive time slices is systematically less than 0.005$\pm$0.01. The absolute variation after 10 time slices is less than 0.03$\pm$0.04. Therefore the inferred parameters do not vary abruptly over time. We recall that the model used here is deliberately simplistic for demonstration; low metrics do not means the Simple Dynamic prior does not work, but instead that it should be coupled to a more complex model (any of SIMSBM(x,y,...)).

\subsubsection{Case study}
As an illustration of what SDSBM has to offer, we plot in Fig.~\ref{fig-epigraphy} a possible visualization of the memberships evolution in time for the epigraphy dataset. On the left and on the right, we show the items that are considered in the visualization. The time goes from left (100BC) to right (500AD), and the flows represent the membership transfers between epochs. The grey bars represent the clusters. We manually annotated them by looking at their composition.
From this figure, we can recover several historical facts: military presence in Rome was scarce for most of the times considered; Italy concentrates less military presence as time goes (due to its spread over the now extended empire), until the 3rd century crisis that led to its re-militarization; most of the slaves that have been accorded an inscription are located in Italia throughout time; the religious functions (Augusti) are evenly spread on the territory at all times; the libertii (freed slaves) inscriptions are essentially present in Rome and in Italy, etc. Obviously, dedicated works are needed in order to support these illustrative claims, and we believe SDSBM can provide such extended comprehension of these processes.

\section{Conclusion}
We introduced a simple way to model time in dynamic weighted and labeled networks by assuming a dynamic Mixed-Membership SBM. Our method consists in defining the Simple Temporal prior, ready to plug into any of a whole class of existing static MMSBMs. Time is considered under the single assumption that network's ties do not vary abruptly. 

We assessed the performance of the proposed method by defining the SDSBM and testing it in several controlled situations on synthetic datasets. In particular, we show that our prior allows stable performances with respect to the dataset slicing, and that it works well under challenging conditions (small amounts of data or high entropy blocks interaction matrix). We tested SDSBM on large scale real-world datasets and showed it yields better results than two proposed baselines. Finally, we illustrate an application interest on a dataset of Latin inscriptions that indirectly narrates the evolution of the Roman Empire.

We discussed the advantages of using our approach: uneven slicing of observations in time, heterogeneous dynamic time-scales, time-dependent blocks-interaction matrix, and informativeness of the prior; exploring these directions on real-world data may help retrieving meaningful clusters on useful applications. On a further note, we believe that a key interest of our approach is the small amount of data needed to get satisfactory performances. This point is fundamental to a number of social sciences, and we believe our approach could ease the incorporation of automated learning methods in these fields.

\bibliographystyle{ACM-Reference-Format}
\balance
\bibliography{Bibliography}

\end{document}